\documentclass[runningheads]{llncs}
\usepackage[T1]{fontenc}
\usepackage{graphicx}
\usepackage{amsmath}
\usepackage{amssymb}
\usepackage{booktabs}
\usepackage[table]{xcolor}
\usepackage{pifont}
\usepackage{booktabs}
\usepackage{multirow}
\usepackage{pifont}
\usepackage{marvosym}
\usepackage{hyperref}
\usepackage{color}

\begin{document}
\title{A Zero-shot Generalized Graph Anomaly Detection Framework via Node Reconstruction}
\titlerunning{AlignGAD: A Zero-shot Generalized GAD Framework}
%

\author{Phan Nguyen\textsuperscript{\Letter} \and
Dat Cao \and
Hien Chu \and
Khue Hoang}
\authorrunning{Phan Nguyen et al.}
\institute{School of Computing, KAIST \\ \email{\{nhphan,ctiendat987,hienchuphan,khuehoangcs\}@kaist.ac.kr}}
%
\maketitle              
\begin{abstract}
Cross-domain graph anomaly detection (GAD) aims to identify abnormal nodes in unseen target graphs, showing strong potential in real-world applications with heterogeneous graph data. However, existing methods often depend on dataset-specific feature semantics and structural patterns, which limits their ability to generalize across different domains. To address this challenge, we propose AlignGAD, a zero-shot generalized graph anomaly detection framework. Our framework is built upon three key components: a Global Unification Module that aligns heterogeneous node features and normalizes graph signals in the spectral domain; a Clustering Module that constructs cluster-aware graph views to capture group-level abnormal patterns; and a Node Discrepancy Scoring Module that measures reconstruction discrepancy and aggregates anomaly evidence from different graph views. Experiments on multiple real-world datasets demonstrate the effectiveness of AlignGAD under the zero-shot GAD setting. \footnote{Our code is available at: \href{https://github.com/nhphan/AlignGAD}{https://github.com/nhphan/AlignGAD}}
\end{abstract}

\keywords{Zero-shot Learning \and Graph Anomaly Detection \and Information Unification \and Clustering \and Generalized Framework.}
\section{Introduction}

\begin{figure}
\centering
\includegraphics[width=0.8\textwidth]{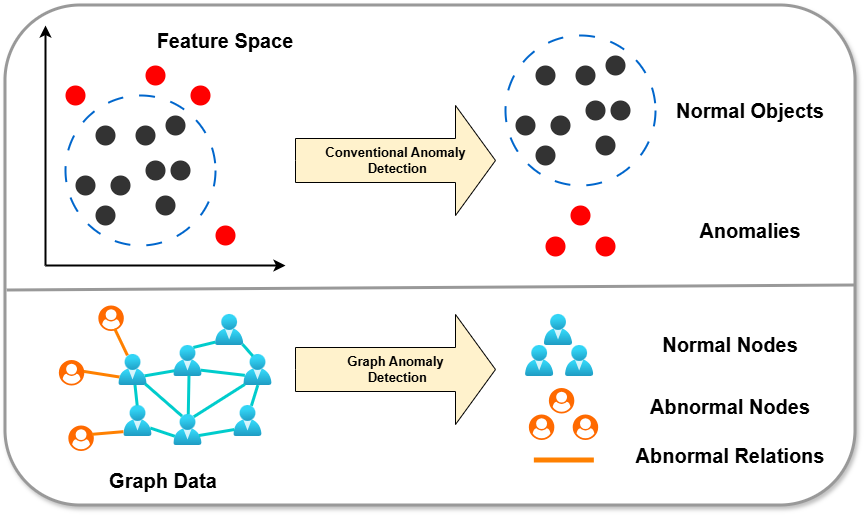}
\caption{Difference between conventional anomaly detection and graph anomaly detection. Conventional anomaly detection focuses on isolated objects in feature space, while graph anomaly detection considers both node attributes and graph relations to identify abnormal nodes and abnormal connections.} \label{fig_intro}
\end{figure}

Graph Anomaly Detection (GAD) aims to identify nodes that deviate from normal patterns in graph-structured data. It has become important in many real-world applications, including multimedia security~\cite{FuGLAD,survey_gnn_time}, social network analysis~\cite{RustGraph,sur_distributed}, and intelligent surveillance~\cite{adver_gnn_multi_time,zheng_correlation-aware_2023}. However, most existing GAD methods are designed for a specific graph or domain. They often depend on domain-specific structural and semantic patterns~\cite{Pang_Xiao_Tai_Cheng_Zhou_2024}, which makes them difficult to transfer to unseen graphs with different feature spaces, graph structures, and anomaly behaviors~\cite{lg_fgad}.

Recently, generalized GAD has attracted increasing attention as a more practical setting. Instead of training a separate detector for each dataset, the goal is to train one model on source graphs and directly apply it to unseen target graphs. Some recent studies have explored this direction, such as ARC~\cite{ARC}, but generalizing across heterogeneous graphs remains challenging. First, graphs from different domains may have different feature dimensions and semantic meanings, making direct feature transfer unreliable. Second, anomalies are not always expressed at the individual node level. In many cases, a node may look less suspicious alone, but become more informative when considered together with the cluster or group it belongs to.

To address these challenges, we propose AlignGAD, a zero-shot framework for graph anomaly detection across heterogeneous graphs. AlignGAD first uses a Global Unification Module to align node feature dimensions and normalize graph signals in the spectral domain. It then introduces a Clustering Module to build cluster-aware graph views, allowing the model to capture group-level abnormal patterns that support node-level detection. Finally, a Node Discrepancy Scoring Module reconstructs node features and measures the discrepancy between input and generated representations. The final anomaly score is obtained by aggregating evidence from both node-level and cluster-level views. Our main contributions are summarized as follows:
\begin{itemize}
    \item We propose AlignGAD, a zero-shot GAD framework that learns from source graphs and directly generalizes to unseen graphs without fine-tuning. By aligning graph features and spectral distributions, AlignGAD reduces the dependence on domain-specific semantics.

    \item  We introduce a cluster-aware discrepancy scoring strategy that combines node reconstruction with cluster-level graph views. This allows the model to capture both individual node deviations and group-level abnormal patterns.

    \item Extensive experiments on multiple real-world graph datasets demonstrate the effectiveness of AlignGAD under the zero-shot setting. The results show that our framework provides a practical and generalizable solution for anomaly detection across heterogeneous graphs.
\end{itemize}
\section{Related Works}

\begin{figure}[h]
\includegraphics[width=\textwidth]{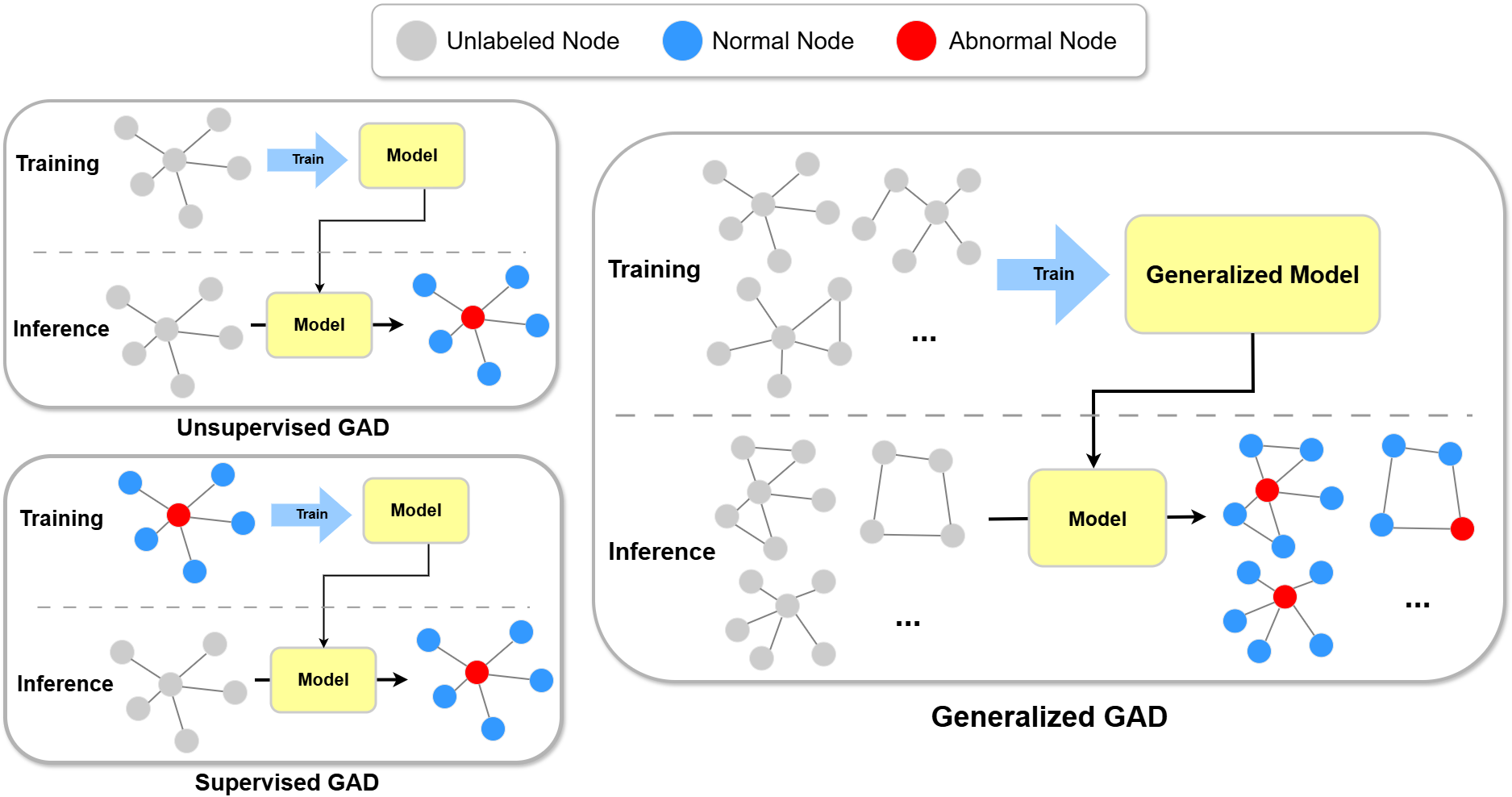}
\caption{Comparison of GAD settings. Conventional supervised and unsupervised GAD methods train and infer within the same domain, while generalized GAD trains on source graphs and directly detects anomalies in unseen target graphs.} \label{fig_gad}
\end{figure}

\paragraph{\textbf{Graph Anomaly Detection.}}
Graph Anomaly Detection (GAD) aims to identify nodes or local graph structures that exhibit abnormal behaviors compared to the majority of the graph \cite{3,23,24}. Existing approaches are generally categorized according to the level of supervision available during training. Supervised methods, such as BGNN \cite{25}, BWGNN \cite{3}, GHRN \cite{GHRN}, and CAGAD \cite{27}, learn discriminative decision boundaries from labeled normal and anomalous nodes, but their applicability is often limited by the scarcity of anomaly annotations. Semi-supervised approaches alleviate this issue by assuming that only a subset of normal nodes is available, as demonstrated by S-GAD \cite{28}, which synthesizes artificial outliers from normal samples to train a one-class classifier. In contrast, unsupervised methods operate without any labels and typically rely on reconstruction or representation learning objectives. For example, DOMINANT \cite{dominant} employs a graph autoencoder to reconstruct graph topology and node attributes, using reconstruction errors as anomaly scores, while CoLA \cite{16} leverages contrastive learning to capture discrepancies between target nodes and their neighborhood subgraphs. Methods such as TAM \cite{TAM} and GCTAM \cite{30} further extend one-class learning paradigms to graph data by optimizing affinity-based objectives in the latent space. Although these methods achieve strong performance on individual datasets, they generally require retraining for each new graph, limiting their ability to generalize across domains.

\paragraph{\textbf{Generalist Graph Anomaly Detection.}}
To overcome the limited transferability of conventional GAD models, recent studies have explored generalist graph anomaly detection frameworks that seek to learn reusable anomaly knowledge across diverse graph domains. While some cross-domain approaches \cite{31,32} can adapt to unseen datasets, they often depend on strong correlations between source and target graphs, such as aligned feature spaces or consistent semantic meanings. ARC \cite{ARC} introduces a unified GAD framework based on in-context learning, where node representations from different graphs are projected into a shared latent space, residual neighborhood information is encoded through an ego-neighbor graph encoder, and anomaly scores are computed via cross-attention against a small set of normal reference samples. This design enables adaptation to new graphs with minimal additional supervision. More recently, UNPrompt \cite{UNPrompt} proposes a zero-shot generalist GAD framework trained on a single source graph. By performing coordinate-wise normalization across node attributes and learning transferable neighborhood prompts, UNPrompt estimates anomaly scores through the predictability of latent node representations. Despite their promising results, both methods exhibit notable limitations: ARC still requires a few normal target-domain samples during inference and may not fully eliminate domain shifts, whereas UNPrompt relies on consistent attribute semantics and a single-source training paradigm, which can reduce its robustness when graph structures and feature distributions vary substantially across domains.
\section{Preliminary}

\paragraph{\textbf{Notations.}} The structure of $i$-th graph is defined as $\mathcal{G}_i = (\mathcal{V}_i, \mathcal{E}_i, A_i, X_i)$. Here, $\mathcal{V}_i = \{v_1,...,v_{N_i}\}$ is the set of nodes with size $N_i = |\mathcal{V}_i|$, and $\mathcal{E}_i \subseteq \mathcal{V}_i\times\mathcal{V}_i$ represents the edge set. The structural information is captured by the adjacency matrix $A_i \in \{0,1\}^{N_i\times N_i}$. The node feature matrix $X_i \in \mathbb{R}^{N_i \times d_i}$ encodes node attributes, with $d_i$ denoting the feature dimension.

\paragraph{\textbf{Conventional GAD.}} 
Given a graph $\mathcal{G}_i$, conventional GAD aims to learn a scoring function $f: \mathcal{V}_i \rightarrow \mathbb{R}$ that assigns an anomaly score to each node. Let $\mathcal{V}_i^{n}$ and $\mathcal{V}_i^{a}$ denote the normal and anomalous node sets, respectively, where anomalous nodes are usually much fewer than normal nodes, i.e., $|\mathcal{V}_i^{a}| \ll |\mathcal{V}_i^{n}|$. The scoring function is expected to give higher scores to anomalous nodes than normal ones, namely $f(v_a) > f(v_n)$ for $v_a \in \mathcal{V}_i^{a}$ and $v_n \in \mathcal{V}_i^{n}$. Most conventional GAD methods train and test within the same graph or domain, which makes them sensitive to dataset-specific feature semantics and structural patterns.

\paragraph{\textbf{Zero-shot GAD.}} 
Zero-shot GAD considers a more practical but harder setting, where a model is trained on source graphs $\mathcal{G}_{\mathrm{train}} = \{\mathcal{G}_1, \ldots, \mathcal{G}_m\}$ and directly applied to unseen target graphs $\mathcal{G}_{\mathrm{test}} = \{\mathcal{G}_{m+1}, \ldots, \mathcal{G}_{m+t}\}$ without further tuning. During inference, no target-domain labels or additional supervision are available. The goal is to learn a general scoring function $f_{\theta}$ that can assign reliable anomaly scores $s_v = f_{\theta}(v, \mathcal{G}_j)$ for nodes in any unseen graph $\mathcal{G}_j \in \mathcal{G}_{\mathrm{test}}$. This setting requires the model to capture transferable graph patterns rather than relying on domain-specific semantics.

\paragraph{\textbf{Graph Fourier Transform.}} 
Graph Fourier Transform (GFT) \cite{GFT} maps graph signals into the spectral domain using the eigendecomposition of the graph Laplacian. For the $i$-th graph, we define the Laplacian as $L_i = D_i - A_i$, where $D_i$ is the degree matrix and $A_i$ is the adjacency matrix. The eigendecomposition of $L_i$ is given by:
\begin{equation}
    L_i = U_i \Lambda_i U_i^{\top}, \quad 0 = \lambda_1 \leq \cdots \leq \lambda_{N_i} \leq 2,
\end{equation}
where $U_i$ contains the orthonormal eigenvectors and $\Lambda_i$ is the diagonal matrix of eigenvalues. Given a graph signal $x \in \mathbb{R}^{N_i}$, the GFT and inverse GFT are defined as $\hat{x}=U_i^{\top}x$ and $x=U_i\hat{x}$, respectively. In the spectral domain, low-frequency components with small $\lambda_k$ usually describe smooth global patterns, while high-frequency components with large $\lambda_k$ capture sharper local variations. This relation can be expressed through the signal smoothness:
\begin{equation}
    x^{\top}L_i x
    = \sum_{k=1}^{N_i} \lambda_k |\hat{x}_k|^2
    = \sum_{k=1}^{N_i} \lambda_k E(\lambda_k).
\end{equation}
A smooth graph signal has most of its energy concentrated in low-frequency components, whereas anomalous or irregular nodes often disturb this pattern by increasing high-frequency energy.

\section{Method}

\begin{figure}[t]
\includegraphics[width=\textwidth]{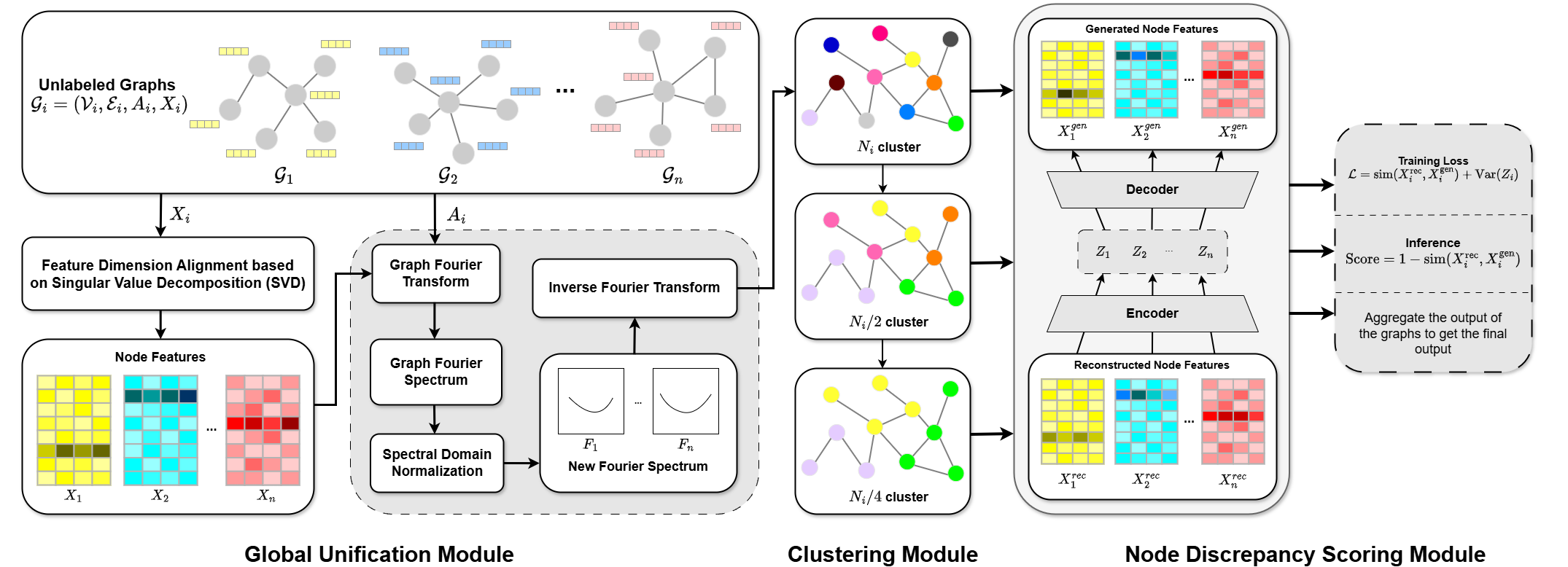}
\caption{The overall framework of AlignGAD. The framework consists of three modules: Global Unification Module, Clustering Module, and Node Discrepancy Scoring Module. The anomaly score is obtained by aggregating the outputs from different graph views.} \label{fig_main_structure}
\end{figure}

In this section, we present AlignGAD, a zero-shot framework for graph anomaly detection across heterogeneous graphs. As shown in Figure~\ref{fig_main_structure}, AlignGAD is built with three main modules. The Global Unification Module first aligns node features from different graphs and normalizes their spectral representations. The Clustering Module then groups nodes into cluster-aware views, allowing the model to capture group-level abnormal patterns that support node-level anomaly prediction. Finally, the Node Discrepancy Scoring Module reconstructs node features and measures the discrepancy between reconstructed and generated representations to produce anomaly scores. The outputs from different graph views are aggregated to obtain the final prediction.

\subsection{Global Unification Module}

A central difficulty in zero-shot GAD is that graphs from different domains often have incompatible feature spaces and different spectral distributions. To make these graphs comparable, the Global Unification Module first projects node features into a shared feature dimension using Singular Value Decomposition (SVD). It then applies Graph Fourier Transform and spectral normalization to reduce distribution gaps in the frequency domain. This module provides a unified representation for heterogeneous graphs.

\paragraph{\textbf{Feature Dimension Alignment.}}
Graphs from different domains often contain node features with inconsistent dimensions, making direct cross-domain learning difficult. Simple operations such as zero-padding or feature truncation may destroy important semantic information and reduce generalization ability. To address this issue, we first align all node features into a shared latent dimension using Singular Value Decomposition (SVD). Given a graph $\mathcal{G}_i$ with node feature matrix $X_i \in \mathbb{R}^{N_i \times d_i}$, we compute
\begin{equation}
    X_i = U_i \Sigma_i V_i^{\top},
\end{equation}
where $U_i$ and $V_i$ are orthogonal matrices, and $\Sigma_i$ contains the singular values. We retain the top-$d'$ singular components to construct a unified representation:
\begin{equation}
    X_i^{align}
    =
    X_i V_i^{(:,1:d')}
    =
    U_i^{(:,1:d')}
    \Sigma_i^{(1:d',1:d')},
\end{equation}
where $d'$ denotes the shared feature dimension across all graphs. In this way, the aligned representation preserves the dominant structural information while reducing feature inconsistencies between heterogeneous graphs.

\paragraph{\textbf{Spectral Domain Normalization.}}
After feature dimension alignment, graphs may still differ in how their signals are distributed over the graph spectrum. Some graphs have smoother signals dominated by low-frequency components, while others contain stronger local variations in high-frequency components. To reduce this spectral gap, we normalize the aligned node features in the frequency domain. For each graph $\mathcal{G}_i$, we first decompose its Laplacian matrix as
\begin{equation}
    L_i = U_i \Lambda_i U_i^{\top},
\end{equation}
where $U_i$ is the graph Fourier basis and $\Lambda_i$ contains the eigenvalues. We then project the aligned features $X_i^{align}$ into the spectral domain:
\begin{equation}
    F_i = U_i^{\top} X_i^{align}
\end{equation}
To make spectral representations more comparable across graphs, we standardize each frequency component:
\begin{equation}
    \hat{F}_i = \frac{F_i - \mu_i}{\sigma_i},
\end{equation}
where the mean and standard deviation are computed as
\begin{equation}
    \mu_i = \frac{1}{N_i}\mathbf{1}^{\top}F_i,
    \qquad
    \sigma_i = \sqrt{\frac{1}{N_i}\mathbf{1}^{\top}(F_i \odot F_i) - \mu_i \odot \mu_i}
\end{equation}
Here, $\mathbf{1} \in \mathbb{R}^{N_i}$ is an all-one vector, and $\odot$ denotes element-wise multiplication. The normalized spectral features are finally mapped back to the node space through inverse GFT:
\begin{equation}
    X_i^{rec} = U_i \hat{F}_i
    = U_i \left( \frac{U_i^{\top}X_i^{align} - \mu_i}{\sigma_i} \right)
\end{equation}
The resulting representation $X_i^{rec}$ preserves the graph structure while reducing spectral distribution differences across domains. This provides a more stable input for the following clustering and node discrepancy scoring modules.

\subsection{Clustering Module}

The Clustering Module is designed to provide group-level structural context for node anomaly detection. Instead of only scoring each node in the original graph, we build a set of cluster-aware graph views with progressively coarser structures. Given the unified node representation $X_i^{rec}$ from the Global Unification Module, we first keep the original graph as the finest view, where each node can be regarded as one cluster. We denote this view as
$\mathcal{G}_i^{(0)} = (\mathcal{V}_i^{(0)}, \mathcal{E}_i^{(0)}, A_i^{(0)}, X_i^{(0)})$, where $X_i^{(0)} = X_i^{rec}$ and $A_i^{(0)} = A_i$.

For the next views, we apply K-means clustering to group nodes into fewer cluster nodes. Specifically, we construct three graph views with cluster sizes $N_i$, $N_i/2$, and $N_i/4$. For the $r$-th clustering view, K-means assigns each node in the previous view to one cluster:
\begin{equation}
    \pi_i^{(r)} = \mathrm{KMeans}(X_i^{(r-1)}, K_i^{(r)}),
    \quad K_i^{(r)} = \left\lfloor \frac{N_i}{2^r} \right\rfloor,
    \quad r \in \{1,2\}.
\end{equation}
Here, $\pi_i^{(r)}$ is the cluster assignment function, and $K_i^{(r)}$ is the number of clusters in the $r$-th view. Each cluster is then treated as a new node, whose feature is computed by averaging the features of all nodes assigned to that cluster:
\begin{equation}
    x_{i,c}^{(r)}
    =
    \frac{1}{|\mathcal{C}_{i,c}^{(r)}|}
    \sum_{v \in \mathcal{C}_{i,c}^{(r)}} x_{i,v}^{(r-1)},
\end{equation}
where $\mathcal{C}_{i,c}^{(r)}$ denotes the set of nodes assigned to the $c$-th cluster in view $r$. The adjacency matrix of the clustered graph is also updated according to the connections between clusters:
\begin{equation}
    A_{i,cd}^{(r)} =
    \mathbb{I}
    \left[
    \exists u \in \mathcal{C}_{i,c}^{(r)}, 
    \exists v \in \mathcal{C}_{i,d}^{(r)}
    \text{ such that }
    A_{i,uv}^{(r-1)} = 1
    \right].
\end{equation}
In this way, the second view is built from the nodes in the first view, and the third view is built from the cluster nodes in the second view. These progressively coarsened graphs allow the model to observe whether a node is suspicious not only by itself, but also through the behavior of the group it belongs to. The resulting graph views are then passed to the Node Discrepancy Scoring Module for anomaly scoring.

\subsection{Node Discrepancy Scoring Module}

The Node Discrepancy Scoring Module measures how much a node deviates from the representation generated by a graph autoencoder. After the Clustering Module, each graph $\mathcal{G}_i$ is represented by three graph views, denoted as $\{\mathcal{G}_i^{(r)}\}_{r=0}^{2}$. Here, $\mathcal{G}_i^{(0)}$ is the finest view with $N_i$ nodes, while $\mathcal{G}_i^{(1)}$ and $\mathcal{G}_i^{(2)}$ are the coarser cluster views with approximately $N_i/2$ and $N_i/4$ cluster nodes. For each view $\mathcal{G}_i^{(r)} = (\mathcal{V}_i^{(r)}, \mathcal{E}_i^{(r)}, A_i^{(r)}, X_i^{(r)})$, we use the same encoder-decoder structure to reconstruct node features and compute discrepancy scores.

The encoder aggregates neighborhood information and maps the input node features into latent representations. For the $l$-th encoder layer, we define
\begin{equation}
    H_i^{(r,l+1)}
    =
    \sigma\left(
    \hat{A}_i^{(r)} H_i^{(r,l)} W^{(l)}
    \right)
    +
    R\left(H_i^{(r,l)}\right),
\end{equation}
where $H_i^{(r,0)} = X_i^{(r)}$, $\hat{A}_i^{(r)}$ is the normalized adjacency matrix of the $r$-th view, $W^{(l)}$ is the learnable weight matrix, $\sigma(\cdot)$ is the activation function, and $R(\cdot)$ denotes the residual connection. The encoder output is denoted as $Z_i^{(r)}$. The decoder then generates node features from this latent representation:
\begin{equation}
    X_i^{gen,(r)}
    =
    \mathrm{Decoder}
    \left(
    Z_i^{(r)}, A_i^{(r)}
    \right)
\end{equation}
If a node follows the common structural pattern of its neighborhood, its generated representation is expected to stay close to the input representation. In contrast, anomalous nodes or anomalous clusters tend to produce larger reconstruction discrepancies.

During training, we optimize the autoencoder over all three graph views. The objective contains a reconstruction discrepancy term and a latent variance regularization term:
\begin{equation}
    \mathcal{L}
    =
    \frac{1}{3}
    \sum_{r=0}^{2}
    \left[
    \mathbb{E}
    \left(
    1 -
    \cos
    \left(
    X_i^{(r)}, X_i^{gen,(r)}
    \right)
    \right)^{\alpha}
    +
    \beta
    \mathcal{L}_{var}^{(r)}
    \right],
    \label{eq:loss}
\end{equation}
where $\alpha > 1$ controls the sensitivity of the reconstruction discrepancy, and $\beta > 0$ controls the strength of the variance regularization. The variance loss is defined as
\begin{equation}
    \mathcal{L}_{var}^{(r)}
    =
    \frac{1}{N_i^{(r)} d'}
    \sum_{n=1}^{N_i^{(r)}}
    \sum_{d=1}^{d'}
    \left(
    Z_{i,n,d}^{(r)}
    -
    \frac{1}{N_i^{(r)}}
    \sum_{n'=1}^{N_i^{(r)}}
    Z_{i,n',d}^{(r)}
    \right)^2,
\end{equation}
where $N_i^{(r)} = |\mathcal{V}_i^{(r)}|$ is the number of nodes in the $r$-th graph view, $d'$ is the unified feature dimension, and $Z_{i,n,d}^{(r)}$ is the $d$-th latent feature of node $n$. This regularization encourages the latent representations to remain stable and reduces the influence of noisy local variations.

At inference time, we compute an anomaly score for each node in each graph view by measuring the cosine distance between its input feature and generated feature:
\begin{equation}
    s_i^{(r)}(u)
    =
    1
    -
    \cos
    \left(
    x_{i,u}^{(r)}, x_{i,u}^{gen,(r)}
    \right)
    =
    1
    -
    \frac{
    x_{i,u}^{(r)}
    \cdot
    x_{i,u}^{gen,(r)}
    }{
    \left\|
    x_{i,u}^{(r)}
    \right\|
    \left\|
    x_{i,u}^{gen,(r)}
    \right\|
    }
\end{equation}
Here, $x_{i,u}^{(r)}$ and $x_{i,u}^{gen,(r)}$ denote the input and generated features of node $u$ in view $r$, respectively. A larger value of $s_i^{(r)}(u)$ indicates a stronger discrepancy and therefore a higher anomaly likelihood.

Since the coarser graph views contain cluster nodes, their scores are mapped back to the original nodes. Let $c_r(v)$ denote the node or cluster in view $r$ that contains the original node $v \in \mathcal{V}_i$. The final anomaly score of $v$ is computed by max aggregation:
\begin{equation}
    S_i(v)
    =
    \max_{r \in \{0,1,2\}}
    s_i^{(r)}
    \left(
    c_r(v)
    \right)
\end{equation}
This aggregation keeps the strongest anomaly evidence across different views. Therefore, a node can receive a high final score either because it is individually abnormal in the finest view, or because it belongs to a suspicious cluster in a coarser view. During evaluation, $S_i(v)$ is compared with the original node-level ground-truth label, so no additional cluster-level labels are required.
\section{Experiments}

\subsection{Experimental Setup}

\paragraph{\textbf{Datasets.}} We use twelve datasets from different distributions and domains. The training datasets include ACM, BlogCatalog, Flickr, and Facebook. The test datasets include Cora, CiteSeer, PubMed, Reddit, CS, Photo, Amazon, and YelpChi. More details about the datasets are in Table~\ref{tab:dataset_stat}.

\newcommand{\cmark}{\ding{51}}

\begin{table}
\centering
\caption{The statistics of datasets.}
\label{tab:dataset_stat}
\small
\setlength{\tabcolsep}{5pt}
\begin{tabular}{l|cc|cccc}
\toprule
Dataset & Train & Test & \#Nodes & \#Edges & \#Features & \%Anomaly \\
\midrule
Cora        & -      & \cmark & 2,708  & 5,429   & 1,433  & 5.53 \\
CiteSeer    & -      & \cmark & 3,327  & 4,732   & 3,703  & 4.50 \\
ACM         & \cmark & -      & 16,484 & 71,980  & 8,337  & 3.62 \\
PubMed      & -      & \cmark & 19,717 & 44,338  & 500    & 3.04 \\
BlogCatalog & \cmark & -      & 5,196  & 171,743 & 8,189  & 5.77 \\
Flickr      & \cmark & -      & 7,575  & 239,738 & 12,047 & 5.94 \\
Facebook    & \cmark & -      & 1,081  & 55,104  & 576    & 2.31 \\
Reddit      & -      & \cmark & 10,984 & 168,016 & 64     & 3.33 \\
CS          & -      & \cmark & 18,333 & 81,894  & 6,805  & 3.27 \\
Photo       & -      & \cmark & 7,650  & 119,081 & 745    & 5.88 \\
Amazon      & -      & \cmark & 10,244 & 175,608 & 25     & 6.76 \\
YelpChi     & -      & \cmark & 23,831 & 49,315  & 32     & 5.10 \\
\bottomrule
\end{tabular}
\end{table}

\paragraph{\textbf{Competing Methods.}} We compare AlignGAD with eleven competing GAD methods, including four supervised methods (GCN~\cite{GCN}, GAT~\cite{GAT}, GHRN~\cite{GHRN}, and NRGL~\cite{NRGL}), five unsupervised methods (AnomalyDAE~\cite{anomalydae}, DOMINANT~\cite{dominant}, TAM~\cite{TAM}, GADAM~\cite{GADAM}, and GGAD~\cite{GGAD}), and two generalist methods (UNPrompt~\cite{UNPrompt} and ARC~\cite{ARC}). All methods are trained and evaluated on the same training and test datasets.

\paragraph{\textbf{Experiment Setting.}} All baselines are implemented using their codebases with default optimization and hyperparameter settings under the same zero-shot protocol. In AlignGAD, we use SVD to project node features of all graphs to an 8-dimensional shared space. The node discrepancy model uses a 3-layer GCN encoder and a 2-layer GCN decoder, with a linear transformation layer between them to improve representation capacity. We train the model on the source graphs and directly evaluate it on all unseen target graphs without fine-tuning.

\paragraph{\textbf{Evaluation Metrics.}} We evaluate all methods using Area Under the Receiver Operating Characteristic Curve (AUROC) and Area Under the Precision-Recall Curve (AUPRC). Higher AUROC and AUPRC values indicate better anomaly detection performance.

\subsection{Performance}

\newcommand{\xmark}{\ding{55}}

\begin{table}
\centering
\caption{AUROC and AUPRC results compared with baselines on 8 real-world datasets. For each dataset, the best performance within each metric is boldfaced.}
\label{tab:main_results}
\small
\setlength{\tabcolsep}{4pt}
\setlength{\arrayrulewidth}{0.8pt}
\resizebox{\textwidth}{!}{
\begin{tabular}{c|c|l|cccccccc}
\toprule[1.2pt]
\textbf{Metric} & \textbf{Type} & \textbf{Method} & \textbf{PubMed} & \textbf{CS} & \textbf{CiteSeer} & \textbf{Photo} & \textbf{Amazon} & \textbf{Cora} & \textbf{Reddit} & \textbf{YelpChi} \\
\midrule[1.0pt]

\multirow{12}{*}{\textbf{AUROC}}
& \multirow{4}{*}{\textbf{Supervised}}
& GCN (ICLR'17)       & 0.537 & 0.544 & 0.667 & 0.533 & 0.503 & 0.559 & 0.498 & 0.465 \\
& & GAT (ICLR'18)      & 0.427 & 0.385 & 0.399 & 0.488 & 0.489 & 0.524 & 0.518 & \textbf{0.518} \\
& & GHRN (WebConf'23)  & 0.378 & 0.348 & 0.619 & 0.585 & 0.456 & 0.548 & 0.505 & 0.501 \\
& & NRGL (IJCAI'24)    & 0.362 & 0.568 & 0.573 & 0.491 & 0.439 & 0.562 & 0.470 & 0.474 \\
\cmidrule[0.8pt](lr){2-11}

& \multirow{5}{*}{\textbf{Unsupervised}}
& AnomalyDAE (ICASSP'20) & 0.604 & 0.619 & 0.484 & 0.530 & 0.308 & 0.530 & 0.459 & 0.407 \\
& & DOMINANT (SDM'19)     & 0.391 & 0.500 & 0.543 & 0.390 & 0.497 & 0.548 & 0.484 & 0.473 \\
& & TAM (NeurIPS'23)      & 0.664 & 
\textbf{0.630} & 0.557 & 0.495 & 0.601 & 0.437 & 0.502 & 0.486 \\
& & GADAM (ICLR'24)       & 0.462 & 0.497 & 0.587 & 0.442 & 0.564 & 0.503 & 0.453 & 0.462 \\
& & GGAD (NeurIPS'24)     & 0.289 & 0.412 & 0.252 & 0.498 & 0.388 & 0.295 & 0.519 & 0.513 \\
\cmidrule[0.8pt](lr){2-11}

& \multirow{3}{*}{\textbf{Generalist GAD}}
& UNPrompt (IJCAI'25)    & 0.654 & 0.584 & 0.607 & 0.487 & 0.430 & 0.627 & 0.508 & 0.444 \\
& & ARC (NeurIPS'24)      & 0.557 & 0.629 & 0.574 & 0.463 & 0.456 & 0.613 & 0.481 & 0.482 \\
& & \textbf{AlignGAD (Ours)} & \textbf{0.756} & 0.612 & \textbf{0.699} & \textbf{0.679} & \textbf{0.602} & \textbf{0.630} & \textbf{0.582} & 0.392 \\
\midrule[1.2pt]

\multirow{12}{*}{\textbf{AUPRC}}
& \multirow{4}{*}{\textbf{Supervised}}
& GCN (ICLR'17)       & 0.033 & 0.036 & 0.078 & 0.063 & 0.068 & 0.063 & 0.033 & 0.047 \\
& & GAT (ICLR'18)      & 0.024 & 0.024 & 0.035 & 0.053 & 0.066 & 0.053 & 0.037 & 0.051 \\
& & GHRN (WebConf'23)  & 0.026 & 0.073 & 0.057 & 0.063 & 0.063 & 0.067 & 0.043 & 0.066 \\
& & NRGL (IJCAI'24)    & 0.064 & 0.083 & 0.053 & 0.057 & \textbf{0.115} & 0.044 & 0.038 & \textbf{0.069} \\
\cmidrule[0.8pt](lr){2-11}

& \multirow{5}{*}{\textbf{Unsupervised}}
& AnomalyDAE (ICASSP'20) & 0.045 & 0.184 & 0.040 & 0.077 & 0.044 & 0.056 & 0.030 & 0.042 \\
& & DOMINANT (SDM'19)     & 0.021 & 0.038 & 0.030 & 0.036 & 0.056 & 0.051 & 0.038 & 0.049 \\
& & TAM (NeurIPS'23)      & 0.095 & 0.044 & 0.052 & 0.054 & 0.062 & 0.046 & 0.033 & 0.057 \\
& & GADAM (ICLR'24)       & 0.102 & 0.062 & 0.072 & 0.081 & 0.074 & 0.052 & 0.036 & 0.055 \\
& & GGAD (NeurIPS'24)     & 0.019 & 0.073 & 0.030 & 0.079 & 0.050 & 0.035 & 0.033 & 0.061 \\
\cmidrule[0.8pt](lr){2-11}

& \multirow{3}{*}{\textbf{Generalist GAD}}
& UNPrompt (IJCAI'25)    & 0.074 & 0.048 & 0.062 & 0.057 & 0.017 & 0.083 & 0.034 & 0.043 \\
& & ARC (NeurIPS'24)      & 0.071 & \textbf{0.197} & 0.072 & 0.055 & 0.018 & 0.099 & 0.030 & 0.048 \\
& & \textbf{AlignGAD (Ours)}  & \textbf{0.238} & 0.131 & \textbf{0.159} & \textbf{0.192} & 0.028 & \textbf{0.117} & \textbf{0.045} & 0.038 \\
\bottomrule[1.2pt]
\end{tabular}
}
\setlength{\arrayrulewidth}{0.4pt}
\end{table}

Table~\ref{tab:main_results} presents the AUROC and AUPRC results of AlignGAD compared with supervised, unsupervised, and generalist GAD baselines. Based on the results, we highlight the following observations:

\begin{itemize}
    \item \textbf{AlignGAD achieves strong overall performance in the zero-shot setting.} Without target-domain fine-tuning, AlignGAD obtains the best AUROC on six out of eight datasets and the best AUPRC on five datasets. This shows that the proposed framework can learn anomaly-related patterns that remain useful across unseen graphs.

    \item \textbf{Cluster-aware scoring provides useful anomaly evidence beyond individual nodes.} AlignGAD performs particularly well on PubMed, CiteSeer, Photo, Cora, and Reddit, where it achieves the best result in at least one evaluation metric. This suggests that assigning scores through cluster-level graph views helps reveal abnormal patterns that may be weak at the individual node level.

    \item \textbf{AlignGAD reduces the dependence on dataset-specific supervision.} Supervised baselines can perform well on some target graphs, such as GAT on YelpChi and NRGL on Amazon, but their behavior is less consistent under domain shift. In contrast, AlignGAD relies on feature alignment and reconstruction discrepancy rather than target labels, making it more suitable for the generalized GAD scenario.
\end{itemize}

\subsection{Ablation Study}

To examine the contribution of each component in AlignGAD, we conduct an ablation study on four representative test datasets. We consider three variants: w/o Global Unification, which removes the whole global unification process; w/o Spectral Norm., which keeps feature dimension alignment but removes spectral domain normalization; and w/o Clustering, which uses only the original graph without cluster-aware graph views. As shown in Table~\ref{tab:ablation}, removing the Global Unification Module generally leads to performance degradation on PubMed, CiteSeer, Photo, and Cora, indicating that aligning heterogeneous graph features is important for reducing cross-domain discrepancy. The drop becomes more significant when spectral normalization is removed, especially on PubMed and CiteSeer, showing that frequency-domain normalization helps produce more stable representations across unseen graphs. The w/o Clustering variant also performs worse than the full model on PubMed, CiteSeer, and Photo, which confirms that cluster-aware graph views provide useful group-level information beyond individual node reconstruction. Although w/o Spectral Norm. achieves a slightly higher result on Cora, the full AlignGAD model performs more consistently across datasets. Overall, these results show that global unification, spectral normalization, and clustering play complementary roles in improving zero-shot graph anomaly detection.

\begin{table}[t]
\centering
\caption{Ablation study on the key components of AlignGAD. The term ``w/o'' indicates ``without''.}
\label{tab:ablation}
\small
\setlength{\tabcolsep}{7pt}
\begin{tabular}{l|cccc}
\toprule
\textbf{Variation} & \textbf{PubMed} & \textbf{CiteSeer} & \textbf{Photo} & \textbf{Cora} \\
\midrule
AlignGAD & \textbf{0.7560} & \textbf{0.6990} & \textbf{0.6790} & 0.6300 \\
w/o Global Unification & 0.7499 & 0.6807 & 0.6559 & 0.5887 \\
w/o Spectral Norm. & 0.5882 & 0.5969 & 0.6433 & \textbf{0.6573} \\
w/o Clustering & 0.6819 & 0.6300 & 0.6151 & 0.5984 \\
\bottomrule
\end{tabular}
\end{table}

\subsection{Hyperparameter Analysis}
We further study the effect of two important hyperparameters in AlignGAD: $\alpha$ and $\beta$. As defined in Equation~\ref{eq:loss}, $\alpha$ controls the sensitivity of the reconstruction discrepancy, while $\beta$ determines the contribution of the latent variance regularization. We conduct a sensitivity analysis by varying $\alpha \in \{1,2,3,4\}$ and $\beta \in \{0.25,0.5,0.75,1\}$ while keeping all other settings unchanged.

Overall, $\alpha=2$ and $\beta=0.5$ provide the best performance across most settings. A moderate value of $\alpha$ effectively emphasizes reconstruction discrepancies, whereas larger values can make the model overly sensitive to large reconstruction errors. Similarly, moderate variance regularization helps stabilize the latent representations without overly restricting their variation. We also observe relatively small performance changes within the tested ranges, indicating that AlignGAD is generally robust to the choice of these two hyperparameters. Based on this analysis, we use $\alpha=2$ and $\beta=0.5$ as the default setting in our experiments.

\section{Conclusion}
In this paper, we propose AlignGAD, a zero-shot generalized graph anomaly detection framework for heterogeneous graphs. AlignGAD learns transferable anomaly patterns from multiple source graphs and directly applies them to unseen target graphs without fine-tuning. By combining graph feature alignment, cluster-aware graph views, and reconstruction-based discrepancy scoring, the model captures both individual node deviations and group-level abnormal patterns. Experiments on multiple real-world datasets demonstrate the effectiveness of AlignGAD under the zero-shot GAD setting. In future work, we plan to further improve the clustering strategy and explore more adaptive aggregation mechanisms for diverse graph domains.


%
%
\bibliographystyle{splncs04}
\bibliography{mybibliography}

@inproceedings{GFT,
  author       = {Joan Bruna and
                  Wojciech Zaremba and
                  Arthur Szlam and
                  Yann LeCun},
  editor       = {Yoshua Bengio and
                  Yann LeCun},
  title        = {Spectral Networks and Locally Connected Networks on Graphs},
  booktitle    = {2nd International Conference on Learning Representations, {ICLR} 2014,
                  Banff, AB, Canada, April 14-16, 2014, Conference Track Proceedings},
  year         = {2014},
  url          = {http://arxiv.org/abs/1312.6203},
  bibsource    = {dblp computer science bibliography, https://dblp.org}
}

@inproceedings{GCN,
  author       = {Thomas N. Kipf and
                  Max Welling},
  title        = {Semi-Supervised Classification with Graph Convolutional Networks},
  booktitle    = {5th International Conference on Learning Representations, {ICLR} 2017,
                  Toulon, France, April 24-26, 2017, Conference Track Proceedings},
  publisher    = {OpenReview.net},
  year         = {2017},
  url          = {https://openreview.net/forum?id=SJU4ayYgl},
  bibsource    = {dblp computer science bibliography, https://dblp.org}
}

@inproceedings{GAT,
  author       = {Petar Velickovic and
                  Guillem Cucurull and
                  Arantxa Casanova and
                  Adriana Romero and
                  Pietro Li{\`{o}} and
                  Yoshua Bengio},
  title        = {Graph Attention Networks},
  booktitle    = {6th International Conference on Learning Representations, {ICLR} 2018,
                  Vancouver, BC, Canada, April 30 - May 3, 2018, Conference Track Proceedings},
  publisher    = {OpenReview.net},
  year         = {2018},
  url          = {https://openreview.net/forum?id=rJXMpikCZ},
  bibsource    = {dblp computer science bibliography, https://dblp.org}
}

@inproceedings{GHRN,
author = {Gao, Yuan and Wang, Xiang and He, Xiangnan and Liu, Zhenguang and Feng, Huamin and Zhang, Yongdong},
title = {Addressing Heterophily in Graph Anomaly Detection: A Perspective of Graph Spectrum},
year = {2023},
isbn = {9781450394161},
publisher = {Association for Computing Machinery},
address = {New York, NY, USA},
url = {https://doi.org/10.1145/3543507.3583268},
doi = {10.1145/3543507.3583268},
booktitle = {Proceedings of the ACM Web Conference 2023},
pages = {1528–1538},
numpages = {11},
location = {Austin, TX, USA},
series = {WWW '23}
}

@inproceedings{NRGL,
author = {Wu, Junhang and Hu, Ruimin and Li, Dengshi and Huang, Zijun and Ren, Lingfei and Zang, Yilong},
title = {Robust heterophilic graph learning against label noise for anomaly detection},
year = {2024},
isbn = {978-1-956792-04-1},
url = {https://doi.org/10.24963/ijcai.2024/271},
doi = {10.24963/ijcai.2024/271},
booktitle = {Proceedings of the Thirty-Third International Joint Conference on Artificial Intelligence},
articleno = {271},
numpages = {9},
location = {Jeju, Korea},
series = {IJCAI '24}
}

@inproceedings{anomalydae,
  author       = {Haoyi Fan and
                  Fengbin Zhang and
                  Zuoyong Li},
  title        = {Anomalydae: Dual Autoencoder for Anomaly Detection on Attributed Networks},
  booktitle    = {2020 {IEEE} International Conference on Acoustics, Speech and Signal
                  Processing, {ICASSP} 2020, Barcelona, Spain, May 4-8, 2020},
  pages        = {5685--5689},
  publisher    = {{IEEE}},
  year         = {2020},
  url          = {https://doi.org/10.1109/ICASSP40776.2020.9053387},
  doi          = {10.1109/ICASSP40776.2020.9053387},
  bibsource    = {dblp computer science bibliography, https://dblp.org}
}

@inbook{dominant,
author = {Kaize Ding and Jundong Li and Rohit Bhanushali and Huan Liu},
title = {Deep Anomaly Detection on Attributed Networks},
booktitle = {Proceedings of the 2019 SIAM International Conference on Data Mining (SDM)},
chapter = {},
pages = {594-602},
doi = {10.1137/1.9781611975673.67},
URL = {https://epubs.siam.org/doi/abs/10.1137/1.9781611975673.67},
eprint = {https://epubs.siam.org/doi/pdf/10.1137/1.9781611975673.67}
}

@inproceedings{TAM,
author = {Qiao, Hezhe and Pang, Guansong},
title = {Truncated affinity maximization: one-class homophily modeling for graph anomaly detection},
year = {2023},
publisher = {Curran Associates Inc.},
address = {Red Hook, NY, USA},
booktitle = {Proceedings of the 37th International Conference on Neural Information Processing Systems},
articleno = {2154},
numpages = {23},
location = {New Orleans, LA, USA},
series = {NIPS '23}
}

@inproceedings{GADAM,
title={Boosting Graph Anomaly Detection with Adaptive Message Passing},
author={Jingyan Chen and Guanghui Zhu and Chunfeng Yuan and Yihua Huang},
booktitle={The Twelfth International Conference on Learning Representations},
year={2024},
url={https://openreview.net/forum?id=CanomFZssu}
}

@inproceedings{GGAD,
title={Generative Semi-supervised Graph Anomaly Detection},
author={Hezhe Qiao and Qingsong Wen and Xiaoli Li and Ee-Peng Lim and Guansong Pang},
booktitle={The Thirty-eighth Annual Conference on Neural Information Processing Systems},
year={2024},
url={https://openreview.net/forum?id=zqLAMwVLkt}
}

@inproceedings{UNPrompt,
  title     = {Zero-shot Generalist Graph Anomaly Detection with Unified Neighborhood Prompts},
  author    = {Niu, Chaoxi and Qiao, Hezhe and Chen, Changlu and Chen, Ling and Pang, Guansong},
  booktitle = {Proceedings of the Thirty-Fourth International Joint Conference on
               Artificial Intelligence, {IJCAI-25}},
  publisher = {International Joint Conferences on Artificial Intelligence Organization},
  editor    = {James Kwok},
  pages     = {3226--3234},
  year      = {2025},
  month     = {8},
  note      = {Main Track},
  doi       = {10.24963/ijcai.2025/359},
  url       = {https://doi.org/10.24963/ijcai.2025/359},
}

@inproceedings{ARC,
 author = {Liu, Yixin and Li, Shiyuan and Zheng, Yu and Chen, Qingfeng and Zhang, Chengqi and Pan, Shirui},
 booktitle = {Advances in Neural Information Processing Systems},
 doi = {10.52202/079017-1606},
 editor = {A. Globerson and L. Mackey and D. Belgrave and A. Fan and U. Paquet and J. Tomczak and C. Zhang},
 pages = {50772--50804},
 publisher = {Curran Associates, Inc.},
 title = {ARC: A Generalist Graph Anomaly Detector with In-Context Learning},
 volume = {37},
 year = {2024}
}

@ARTICLE{FuGLAD,
  author={He, Shiming and Li, Genxin and Xie, Kun and Sharma, Pradip Kumar},
  journal={IEEE Transactions on Information Forensics and Security}, 
  title={Fusion Graph Structure Learning-Based Multivariate Time Series Anomaly Detection With Structured Prior Knowledge}, 
  year={2024},
  volume={19},
  number={},
  pages={8760-8772},
  doi={10.1109/TIFS.2024.3459631}
}

@ARTICLE{survey_gnn_time,
author={Jin, Ming and Koh, Huan Yee and Wen, Qingsong and Zambon, Daniele and Alippi, Cesare and Webb, Geoffrey I. and King, Irwin and Pan, Shirui},
journal={ IEEE Transactions on Pattern Analysis \& Machine Intelligence },
title={{ A Survey on Graph Neural Networks for Time Series: Forecasting, Classification, Imputation, and Anomaly Detection }},
year={2024},
volume={46},
number={12},
ISSN={1939-3539},
pages={10466-10485},
doi={10.1109/TPAMI.2024.3443141},
url = {https://doi.ieeecomputersociety.org/10.1109/TPAMI.2024.3443141},
publisher={IEEE Computer Society},
address={Los Alamitos, CA, USA},
month=dec}

@article{adver_gnn_multi_time,
author = {Zheng, Bolong and Ming, Lingfeng and Zeng, Kai and Zhou, Mengtao and Zhang, Xinyong and Ye, Tao and Yang, Bin and Zhou, Xiaofang and Jensen, Christian S.},
title = {Adversarial Graph Neural Network for Multivariate Time Series Anomaly Detection},
year = {2024},
issue_date = {Dec. 2024},
publisher = {IEEE Educational Activities Department},
address = {USA},
volume = {36},
number = {12},
issn = {1041-4347},
url = {https://doi.org/10.1109/TKDE.2024.3419891},
doi = {10.1109/TKDE.2024.3419891},
journal = {IEEE Trans. on Knowl. and Data Eng.},
month = dec,
pages = {7612–7626},
numpages = {15}
}

@article{zheng_correlation-aware_2023,
	title = {Correlation-aware {Spatial}-{Temporal} {Graph} {Learning} for {Multivariate} {Time}-series {Anomaly} {Detection}},
	journal = {IEEE Transactions on Neural Networks and Learning Systems (TNNLS)},
	author = {Zheng, Yu and Koh, Huan Yee and Jin, Ming and Chi, Lianhua and Phan, Khoa T and Pan, Shirui and Chen, Yi-Ping Phoebe and Xiang, Wei},
	year = {2023},
}

@article{RustGraph,
author = {Guo, Jianhao and Tang, Siliang and Li, Juncheng and Pan, Kaihang and Wu, Lingfei},
title = {RustGraph: Robust Anomaly Detection in Dynamic Graphs by Jointly Learning Structural-Temporal Dependency},
year = {2024},
issue_date = {July 2024},
publisher = {IEEE Educational Activities Department},
address = {USA},
volume = {36},
number = {7},
issn = {1041-4347},
url = {https://doi.org/10.1109/TKDE.2023.3328645},
doi = {10.1109/TKDE.2023.3328645},
journal = {IEEE Trans. on Knowl. and Data Eng.},
month = jul,
pages = {3472–3485},
numpages = {14}
}

@article{sur_distributed,
author = {Pazho, Armin Danesh and Noghre, Ghazal Alinezhad and Purkayastha, Arnab A and Vempati, Jagannadh and Martin, Otto and Tabkhi, Hamed},
title = {A Survey of Graph-Based Deep Learning for Anomaly Detection in Distributed Systems},
year = {2024},
issue_date = {Jan. 2024},
publisher = {IEEE Educational Activities Department},
address = {USA},
volume = {36},
number = {1},
issn = {1041-4347},
url = {https://doi.org/10.1109/TKDE.2023.3282898},
doi = {10.1109/TKDE.2023.3282898},
journal = {IEEE Trans. on Knowl. and Data Eng.},
month = jan,
pages = {1–20},
numpages = {20}
}

@article{Pang_Xiao_Tai_Cheng_Zhou_2024, title={Graph Anomaly Detection with Diffusion Model-Based Graph Enhancement (Student Abstract)}, volume={38}, url={https://ojs.aaai.org/index.php/AAAI/article/view/30494}, DOI={10.1609/aaai.v38i21.30494}, number={21}, journal={Proceedings of the AAAI Conference on Artificial Intelligence}, author={Pang, Shikang and Xiao, Chunjing and Tai, Wenxin and Cheng, Zhangtao and Zhou, Fan}, year={2024}, month={Mar.}, pages={23610–23612} }

@inproceedings{lg_fgad,
author = {Cai, Jinyu and Zhang, Yunhe and Fan, Jicong and Ng, See-Kiong},
title = {LG-FGAD: an effective federated graph anomaly detection framework},
year = {2024},
isbn = {978-1-956792-04-1},
url = {https://doi.org/10.24963/ijcai.2024/416},
doi = {10.24963/ijcai.2024/416},
booktitle = {Proceedings of the Thirty-Third International Joint Conference on Artificial Intelligence},
articleno = {416},
numpages = {10},
location = {Jeju, Korea},
series = {IJCAI '24}
}

@ARTICLE{31,
  author={Ding, Kaize and Shu, Kai and Shan, Xuan and Li, Jundong and Liu, Huan},
  journal={IEEE Transactions on Neural Networks and Learning Systems}, 
  title={Cross-Domain Graph Anomaly Detection}, 
  year={2022},
  volume={33},
  number={6},
  pages={2406-2415},
  doi={10.1109/TNNLS.2021.3110982}}

@inproceedings{32,
author = {Wang, Qizhou and Pang, Guansong and Salehi, Mahsa and Buntine, Wray and Leckie, Christopher},
title = {Cross-domain graph anomaly detection via anomaly-aware contrastive alignment},
year = {2023},
isbn = {978-1-57735-880-0},
publisher = {AAAI Press},
url = {https://doi.org/10.1609/aaai.v37i4.25591},
doi = {10.1609/aaai.v37i4.25591},
articleno = {522},
numpages = {9},
series = {AAAI'23/IAAI'23/EAAI'23}
}

@inproceedings{30,
author = {Zhang, Xiong and Peng, Hong and He, Zhenli and Xie, Cheng and Jin, Xin and Jiang, Hua},
title = {GCTAM: global and contextual truncated affinity combined maximization model for unsupervised graph anomaly detection},
year = {2025},
isbn = {978-1-956792-06-5},
url = {https://doi.org/10.24963/ijcai.2025/405},
doi = {10.24963/ijcai.2025/405},
booktitle = {Proceedings of the Thirty-Fourth International Joint Conference on Artificial Intelligence},
articleno = {405},
numpages = {9},
location = {Montreal, Canada},
series = {IJCAI '25}
}

@inproceedings{28,
author = {Qiao, Hezhe and Wen, Qingsong and Li, Xiaoli and Lim, Ee-Peng and Pang, Guansong},
title = {Generative semi-supervised graph anomaly detection},
year = {2024},
isbn = {9798331314385},
publisher = {Curran Associates Inc.},
address = {Red Hook, NY, USA},
booktitle = {Proceedings of the 38th International Conference on Neural Information Processing Systems},
articleno = {152},
numpages = {29},
location = {Vancouver, BC, Canada},
series = {NIPS '24}
}

@misc{27,
      title={Counterfactual Data Augmentation with Denoising Diffusion for Graph Anomaly Detection}, 
      author={Chunjing Xiao and Shikang Pang and Xovee Xu and Xuan Li and Goce Trajcevski and Fan Zhou},
      year={2024},
      eprint={2407.02143},
      archivePrefix={arXiv},
      primaryClass={cs.LG},
      url={https://arxiv.org/abs/2407.02143}, 
}

@misc{25,
      title={Boost then Convolve: Gradient Boosting Meets Graph Neural Networks}, 
      author={Sergei Ivanov and Liudmila Prokhorenkova},
      year={2021},
      eprint={2101.08543},
      archivePrefix={arXiv},
      primaryClass={cs.LG},
      url={https://arxiv.org/abs/2101.08543}, 
}

@inproceedings{24,
author = {Gao, Yuan and Wang, Xiang and He, Xiangnan and Liu, Zhenguang and Feng, Huamin and Zhang, Yongdong},
title = {Alleviating Structural Distribution Shift in Graph Anomaly Detection},
year = {2023},
isbn = {9781450394079},
publisher = {Association for Computing Machinery},
address = {New York, NY, USA},
url = {https://doi.org/10.1145/3539597.3570377},
doi = {10.1145/3539597.3570377},
booktitle = {Proceedings of the Sixteenth ACM International Conference on Web Search and Data Mining},
pages = {357–365},
numpages = {9},
location = {Singapore, Singapore},
series = {WSDM '23}
}

@inproceedings{23,
author = {Luo, Xuexiong and Wu, Jia and Beheshti, Amin and Yang, Jian and Zhang, Xiankun and Wang, Yuan and Xue, Shan},
title = {ComGA: Community-Aware Attributed Graph Anomaly Detection},
year = {2022},
isbn = {9781450391320},
publisher = {Association for Computing Machinery},
address = {New York, NY, USA},
url = {https://doi.org/10.1145/3488560.3498389},
doi = {10.1145/3488560.3498389},
booktitle = {Proceedings of the Fifteenth ACM International Conference on Web Search and Data Mining},
pages = {657–665},
numpages = {9},
location = {Virtual Event, AZ, USA},
series = {WSDM '22}
}

@article{16,
   title={Anomaly Detection on Attributed Networks via Contrastive Self-Supervised Learning},
   volume={33},
   ISSN={2162-2388},
   url={http://dx.doi.org/10.1109/TNNLS.2021.3068344},
   DOI={10.1109/tnnls.2021.3068344},
   number={6},
   journal={IEEE Transactions on Neural Networks and Learning Systems},
   publisher={Institute of Electrical and Electronics Engineers (IEEE)},
   author={Liu, Yixin and Li, Zhao and Pan, Shirui and Gong, Chen and Zhou, Chuan and Karypis, George},
   year={2022},
   month=June, pages={2378–2392} }

@InProceedings{3,
  title = 	 {Rethinking Graph Neural Networks for Anomaly Detection},
  author =       {Tang, Jianheng and Li, Jiajin and Gao, Ziqi and Li, Jia},
  booktitle = 	 {International Conference on Machine Learning},
  year = 	 {2022},
}
%




\end{document}